\RequirePackage{fix-cm}
\documentclass[smallextended]{svjour3}       
\smartqed  
\usepackage{graphicx}
\begin{document}

\title{Select Good Regions for Deblurring based on Convolutional Neural Networks \thanks{This work is supported by Youth Innovation Promotion Association No.2020220.}
}


\author{Hang Yang         \and
        Xiaotian Wu       \and
        Xinglong Sun 
}


\institute{H. Yang \at
              Changchun Institute of Optics, Fine Mechanics and Physics, Chinese Academy of Science,
Changchun 130033, China.\\
              Tel.: +86-0431-86176563\\
              Fax: +86-0431-86176399\\
              \email{yanghang@ciomp.ac.cn}           
           \and
           X. Wu \at
              Changchun University of Science and Technology,Changchun 130022, China.
           \and
              X. Sun \at
              Changchun Institute of Optics, Fine Mechanics and Physics, Chinese Academy of Science,
Changchun 130033, China.
}

\date{Received: date / Accepted: date}

\maketitle

\begin{abstract}
The goal of blind image deblurring is to recover sharp image
from one input blurred image with an unknown blur kernel.
Most of image deblurring approaches focus on developing image priors, however, there is not enough attention
to the influence of image details and structures on the blur kernel estimation.
What is the useful image structure and how to choose a good deblurring region?
In this work, we propose a deep neural network model method for selecting good regions to estimate blur kernel.
First we construct image patches with labels and train a deep neural networks, then the learned model is applied to
determine which region of the image is most suitable to deblur.
Experimental results illustrate that the proposed approach is effective, and could be able to select good regions for image deblurring.
\end{abstract}

\section{Introduction}
\label{intro}
Image deblurring has been known in the field of computer vision.
All kinds of imaging equipment, such as hand-held cameras and surveillance cameras,
often appear unwanted blur when capturing pictures.
A motion blurred image is produced as the scene or camera moved during the exposure time.

Under the assumption of spatially-invariant blur across the sensor, we consider it as uniform blur.
The blurring processing is often modelled as the convolution of the latent image $L$ with a blur kernel $k$:
\begin{equation}\label{1}
    B = k\ast L + \gamma
\end{equation}
where $B$ and $\gamma$ denote blurry image and the additive noise, respectively, and "$\ast$" is the convolution operator.
The image deblurring algorithms try to estimate both  $L$ and $k$ simultaneously.
This is a mathematically ill-posed problem, since different pairs of $L$ and $k$ can produce the same $B$.

An effective method of image deblurring is to estimate the blur kernel first,
and then, the deblurring problem is transformed into a non blind deconvolution problem.
In this case, the accuracy of blur kernel estimation directly affects the performance of restoration results.

In order to remove the blur from a single image,
it is necessary to take full use of the information of the blurred image.
However, not all pixels of the blurry image have a positive effect on blur kernel estimation.
For example, the contribution of smooth regions with the estimation of the blur kernel is not much.
In this work, we will focus on what kinds of features of blurry image are useful for blur kernel estimation.
When one selects a good region to estimate the blur kernel, an accurate blur kernel can be obtained, then it can be used to
restore a sharp latent image with high-quality visuals.

In \cite{Levin2009,Fergus2006}, the experimental results show that the region with strong edges can produce better deblurring results.
A variety of gradient based methods have been proposed in \cite{Xu2010,Joshi2008,Cho2010,Shan2008}, which show that
salient edges  with special gradient patterns can favor blur kernel estimation.
In addition, based on the one-dimensional signal examples, it is proved that the edge with short length will have an adverse effect on deblurring \cite{Xu2010}.

In other words, if the whole image is used for image deblurring rather than deliberately selecting good regions,
it is likely to obtain a poor quality result.
Even the fast deblurring approach \cite{Cho2009} is not the best choice for estimating the blur kernel with the whole image
\cite{HuZhe2015}. For this reason and the computational efficiency issue, it is preferable
to select a region, rather than the entire image, for estimating the blur kernel \cite{HuZhe2015}.

\begin{figure*}[h]
  \hspace{-0.65cm}
  \includegraphics[width=1.1\linewidth]{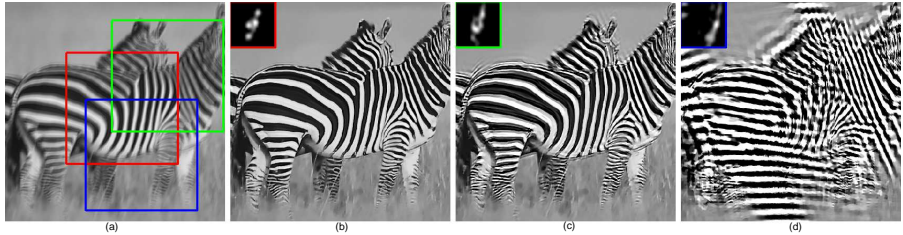}\\
  \vspace{-0.5cm}
  \caption{ Different patches lead to different kernel estimations and
deblurred results. (a)The input blurred image with three sub-windows selected for estimating kernels. (b),(c) and (d)The three images are the recovered images and estimated kernels from these three subwindows using \cite{Xu2010}}\label{horse}
\end{figure*}

In Figure \ref{horse}, we show different kernel estimation results for different regions, and corresponding  different recovered  images. It can be found that there are many difference between them. It is difficult to solve this problem by manual selection or visual inspection of the results. Since the strong edge could not improve blur kernel estimation when the kernel scale is relatively larger than the object \cite{Xu2010}. In addition, there continues to be no answer to the question of which regions or image structures are essential for accurate blur kernel estimation.

In this work, we propose an effective and efficient blur kernel estimation approach based on deep convolution neural networks (CNNs).
Different from empirical understanding and prior knowledge to select good image features, we use CNNs to learn for this task based on
a proposed  dataset with labels. To construct the dataset, we employ the blur kernel acquisition technique introduced in \cite{Levin2009}, and collect a large variability of images.
The labels are generated using the kernel similarity measure \cite{HuZhe2015}.
In the experiments, we demonstrate that the proposed method provides competitive results for selecting appropriate regions on deblurring tasks.

\section{Related Work}

The essence of image deblurring is to estimate an accurate  blur kernel. The main blur kernel estimation methods can be divided into two categories : optimization-based and learning-based methods.
The optimization-based methods focus on exploring image priors for deblurring. Fergus $et~al.$ \cite{Fergus2006} present the sparse gradients prior of natural images and sparse characteristic of blur kernels.
Cho and Lee \cite{Cho2009} utilize image gradient in a multi-scale framework for the deblurring process.
Xu $et~al.$ \cite{Xu2010} propose a two-phase algorithm to improve the blur kernel estimation.
Levin $et~al.$ [14] estimate the blur kernel by optimizing the maximum a posteriori (MAP) model.
Gong $et~al.$ [5] utilize a gradient activation approach to choose a subset of gradients for kernel estimation instead of using the strong edges.
These image deblurring methods are effective for natural blurry images, however they will encounter setbacks when it comes to special types, such as low-illumination, text and human face images.

Numbers of specific image priors have also been proposed to solve this problem \cite{Lai2015,Pan2014,Pan2016,Yan2017,Chen2019}.
Lai $et~al.$ \cite{Lai2015} recover the sharp edges using a color-line prior.
For text image deblurring, Pan $et~al.$ \cite{Pan2014} introduce the L0 regularized prior on both image intensity and gradients.
Recently, dark channel prior for deblurring images is proposed in \cite{Pan2016}, which perform well for low-illumination, text and human face images. Nevertheless,
the dark channel prior is helpless to estimate the blur kernel when the images are bright pixels dominant.
Yan  $et~al.$  \cite{Yan2017} further present an extreme channel prior which incorporates dark channel prior and bright channel prior
channel, it can improve the robustness of the deblurring methods. The Local Maximum Gradient (LMG) prior is presented in \cite{Chen2019}, which
can handle various specific scenarios.

With the success of deep learning in the field of high-level vision problems, there are many CNNs based methods for image deblurring problem.
Schuler $et~al.$ \cite{Schuler2016} estimate blur kernels with a trained deep network and then restore the clear image using a conventional non-blind deconvolution method.
Hradis $et~al.$ \cite{Hradis2015} train a CNNs to reconstruct sharp text images directly without assuming any specific blur models.
Yan $et~al.$ \cite{Yan2016}  propose a deep network and a general regression network to classify the type of the blur kernel and estimate its parameters.
Kupyn $et~al.$ \cite{Kupyn2018,Kupyn2019} introduce two end to end generative adversarial networks (GANs) for  motion deblurring, respectively, it is not needed to estimate blur kernels.
More summary of image restoration is presented in the Ref.\cite{Hosseini2020}.

A large number of works focus on using image prior hypothesis to improve the effect of image deblurring. There are few attention on the influence of image structure features for blur kernel estimation. In this work, our goal is to identify useful image structures for blur kernel estimation and image deblurring.

\section{Learning based Good Regions Selection}
In this section, we describe the motivation of selecting the good image patches, network design, loss function, and implementation details.

\subsection{Motivation}
Many previous works have shown that the smooth region of the image can not provide enough information for the blur kernel estimation,
while regions with many textures may still obtain poor estimation results.
In fact, when the blur motion occurs in a direction similar to the edge, the region full of repeated edges sometimes does not contribute to the problem \cite{HuZhe2015}.

In order to estimate blur kernels, recent algorithms focus on using strong edges or edge distribution [3, 4]. Similar to the problem of texture areas, under appropriate assumptions, sharp edges are of great value for image deblurring. The basic assumption of effective use of sharp edges is that the contrast of the original image maintains the information structure after motion blur. However, not all sharp edges can be effectively used for kernel estimation. Recently, Xu $et~al.$ \cite{Xu2010} show that edges smaller than the blur kernel may be harmful for kernel estimation, and edges with a sufficiently large scale are beneficial for image deblurring. Fergus $et~al.$ \cite{Fergus2006} choose the patch with large variance and low saturation for blur kernel estimation. Bae $et~al.$ \cite{Bae2012} present an approach to select some image regions to estimate the blur kernel based on a pixel-wise measure in terms of non-straightness and edge size. Hu $et~al.$ \cite{HuZhe2015} construct image features using Gabor filter bank,  and learn a binary classifier to select good regions  for deblurring within the Conditional Random Field (CRF) framework. Although the CRF model has achieved good results, it is not suitable for wide application due to the artificially designed image features and its high computational complexity.

Considering these factors, we propose a CNNs based method to learn features from the image blurred region.
In this work, we consider the selecting good regions problem as a binary classification problem,
image regions are classified into positive and negative categories, the positive category means the regions can be used to estimate the blur kernel well, and the negative labeled image region cannot be used to estimate the blur kernel.
Then, we choose the top ranked image patch to estimate the blur kernel for simplicity.
Given the region size, our method can select good region to deblur and does not require manual selection which may lead to incorrect results.

\begin{figure*}[!t]
  \hspace{-0.05cm}
  \includegraphics[width=1.0\linewidth]{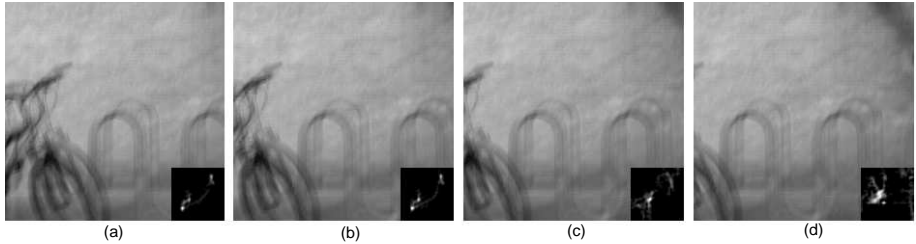}\\
  \vspace{-0.75cm}
  \caption{The estimated kernels from shifting sub-windows. The kernel similarity from (a) to (d) is 0.870,0.864,0.440,0.397, respectively}\label{Shift}
\end{figure*}

To select good image patches for deblurring, we learn the image features by small subwindows.
As we know, two regions with closely overlapping (e.g. shifted by a few pixels in either directions) have similar image features,
However, it should also be pointed out that two neighboring image regions may belong to two different categories.
Figure \ref{Shift} shows an example where we estimate four blur kernels from neighboring four subwindow of size $228 \times 228$.
One can find that although the subwindows appear to be similar, their kernel estimation are very different.
Therefore, when we design the network, we should reduce the translation invariance.

\subsection{ Binary classification network}
Our goal is to train a binary classification convolutional neural networks (CNNs).
The classifier takes an image patch as the input and outputs a single
scalar, which represents the probability of high quality blur kernel can be estimated by using the input patch.

As our aim is to minimize the translation invariance of the network, the CNNs should have fewer pooling layers.
Therefore we improve the ResNet34 \cite{He2016}, and adopt the strided convolution layers instead of pooling layers to improve the classification performance.

\subsection{Loss function}

The input image patch and its label are represented by $x$ and $y$, respectively, and the network parameters are denoted as  $\theta$.
The CNNs learns a mapping function $f(x;\theta) = P(x \in Good\,Region \mid x)$, and predicts the probability $\widehat{y}$ can be estimated by using $x$.

The binary cross entropy loss function is used to optimize the classifier:
\begin{equation}\label{3.1}
    L(\theta) = -\frac{1}{N}\sum_{i=1}^{N} y_{i}\log\widehat{y}_{i} + (1-y_{i})\log(1-\widehat{y}_{i})
\end{equation}
where $N$ represents the number of training samples in a batch, $\widehat{y}_{i}$ =
$f(x;\theta)$ is the output of the network. We assign $y = 1$ for good region to deblur and
$y = 0$ for bad region.

\subsection{Training details}
In this work, we apply the same technique which introduced in \cite{Levin2009} to generate 960 blurred images for training.
First, we choose 30 sharp images including natural, manmade scene, text and face image, and 32 blur kernels with the size
ranging from $11 \times 11 $ to $55 \times 55$. Then, we use the convolution of clear images and blur kernels, and add Gaussian noise to synthesize blurred images, where the variance $\sigma$ of noise is 4.0.

For the model training, we utilize the Stochastic Gradient Descent (SGD)
method for optimizing the network. We use the batch size
of 32, the momentum of 0.9. The learning rate is set to 0.001.
We use pre-training parameters of ResNet34 as initialization, and we fine tune the replaced convolution layer, the third layer, the fourth layer and the full connection layer of the network.
According to our experiments, 20 epochs are enough for convergence.

We train the network parameters using all the 960 blurred images for identifying good
patches to deblur. For each image, we construct a set with overlapping regions of $228\times 228$ pixels
and shifts of 20 pixels. From the experiment, we found that the size of image patches is large enough to obtain the blur kernel estimation when the size of kernel is smaller than $55 \times 55$ pixels.
Given these parameters, for a $450 \times 450$ size image, we can obtain 121 patches as the samples for training.

As the image priori based method has been shown to be robust for the blur kernel estimation, we estimate the blur kernel from each patch using Yang $et~al.$ \cite{Yang2019} during training.

For the blur kernel estimated from the image region, we calculate the kernel similarity proposed in \cite{HuZhe2015} between it and the corresponding ground truth kernel.
If the similarity is larger than the threshold $\lambda$, we label the image region as 1,  otherwise we label the image region as 0. In order to make the proportion of positive and negative samples is close to $1:1$, and  ensure the accuracy of estimated blur kernel, we set the value of the threshold $\lambda = 0.75$.

\section{ Experimental Results}
To verify the superiority of our method, we compare it with three state-of-the-art image blurred regions inference methods \cite{Fergus2006,HuZhe2015,Xu2010} and evaluate the performance using the error metric introduced in \cite{Levin2009} which measures the ratio between recovered error with the estimated blur kernel and recovered error with the ground truth kernel.
The error ratio $ER$ can be written as :
\begin{equation}\label{4.1}
    ER = \frac{\parallel I_{e}-I_{g}\parallel^{2}}{\parallel I_{k_{g}}-I_{g}\parallel^{2}}
\end{equation}
where $I_{e}$ is the estimated image,  $I_{g}$ is the ground truth image, $I_{k_{g}}$ is the deconvolution image with the ground truth kernel $k_{g}$.
The cumulative histogram of recovered error ratio is used to evaluate the effectiveness of the method.

In order to validate whether the trained model can still provide good regions for other deblurring methods,
we apply the approach proposed by Pan $et~al.$ \cite{Pan2016} to estimate the blur kernel and recover the final deconvolution results.

\subsection{Comparison with User-selected Region}
\begin{figure*}[!t]
  \hspace{-1.75cm}
  \includegraphics[width=1.15\linewidth]{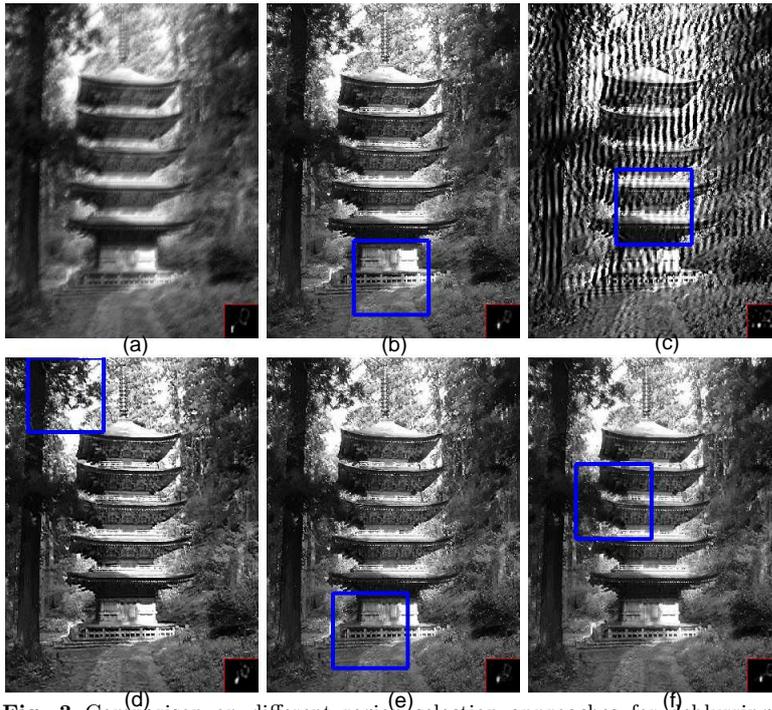}\\
  \vspace{-1.95cm}
  \caption{Comparison on different region selection approaches for deblurring. We use the method
in \cite{Pan2016} for kernel estimation and  non-blind deconvolution. (a) Blurred image. (b)The subwindow (blue box) selected by user. (c) The subwindow (blue box) selected by window selector \cite{Fergus2006} and the deblurred result.  (d)The subwindow (blue box) selected by window selector \cite{Xu2010} and the deblurred result.(e)The subwindow (blue box) selected by window selector \cite{HuZhe2015} and the deblurred result.(f)The subwindow (blue box) selected by  our algorithm and the deblurred result.(Best viewed on a high-resolution display).}\label{tower}
\end{figure*}

In order to prove that our algorithm can replace manual selection, we compare our selected good regions with
user selected patches. Users tend to select the patches with the significant edges, e.g. the subwindows shown in Figure \ref{tower} (b) and Figure \ref{text} (b). User selection strategy works well in some cases, but it usually needs many attempts to obtain good results. On the contrary, the proposed algorithm does not need the guidance of users, and the results of deblurring by the inferred patches of the proposed method outperform that using regions selected by users.

\subsection{Comparison with Region Selection Algorithms}
We compare our method with three other region selection methods for deblurring \cite{Fergus2006,Xu2010,HuZhe2015}.
We implement the region selection algorithm introduced in \cite{Xu2010} that the image restoration process may degrade by
the negative effect of small edges. Automatic regions selector by \cite{HuZhe2015}  does not perform well for saturated
images. In our selector, we learn more effective image features from the training data than \cite{HuZhe2015}.

In this experiment, for simplicity, we select the top ranked patches from the proposed region selection algorithm, although other options may be used. We also use other algorithms to choose a region from each blurred image for estimating the blur kernel, then the state-of-the-art algorithm \cite{Pan2016} is applied to restore the whole image.

\begin{figure*}[!t]
  \hspace{-0.05cm}
  \includegraphics[width=1.0\linewidth]{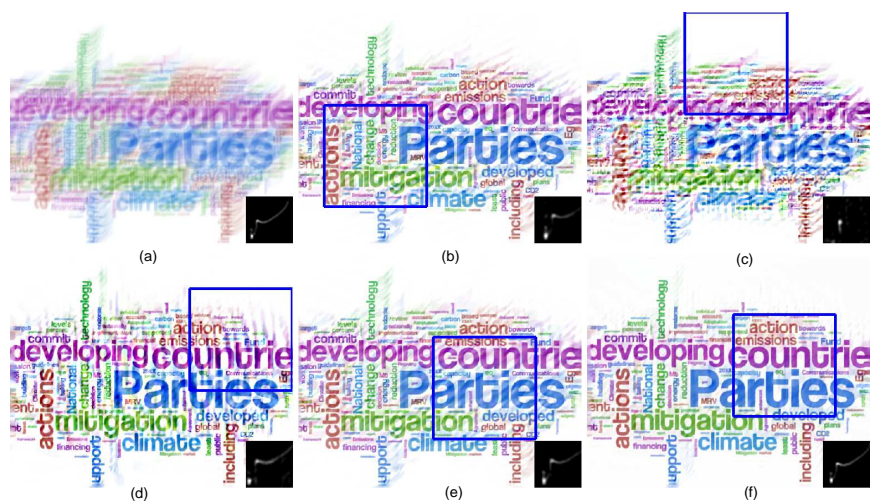}\\
  \vspace{-0.5cm}
  \caption{Comparison on different region selection approaches for deblurring. We use the method
in \cite{Pan2016} for kernel estimation and  non-blind deconvolution. (a) Blurred image. (b)The subwindow (blue box) selected by user. (c) The subwindow (blue box) selected by window selector \cite{Fergus2006} and the deblurred result.  (d)The subwindow (blue box) selected by window selector \cite{Xu2010} and the deblurred result.(e)The subwindow (blue box) selected by window selector \cite{HuZhe2015} and the deblurred result.(f)The subwindow (blue box) selected by our algorithm and the deblurred result.(Best viewed on a high-resolution display).}\label{text}
\end{figure*}

\begin{figure*}[!t]
  \hspace{-1.25cm}
  \includegraphics[width=1.1\linewidth]{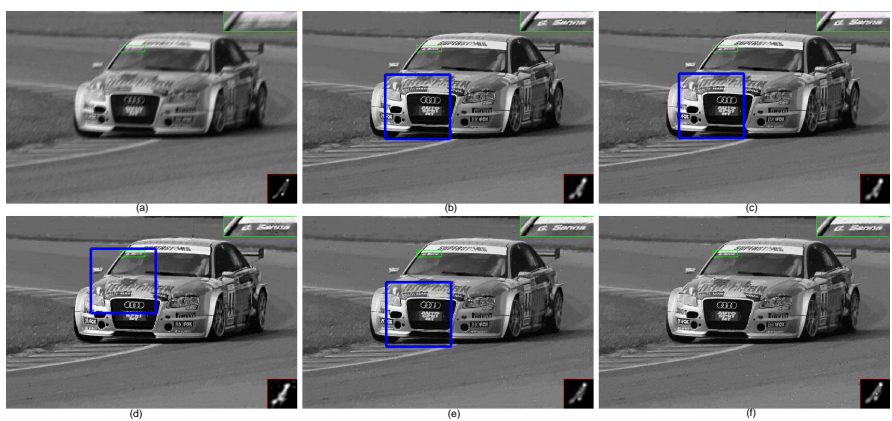}\\
  \vspace{-0.5cm}
  \caption{Comparison on different region selection approaches for deblurring. We use the method
in \cite{Pan2016} for kernel estimation and  non-blind deconvolution. (a) Blurred image. (b)The subwindow (blue box) selected by window selector \cite{Fergus2006} and the deblurred result.  (c)The subwindow (blue box) selected by window selector \cite{Xu2010} and the deblurred result.(d)The subwindow (blue box) selected by window selector \cite{HuZhe2015} and the deblurred result.(e)The subwindow (blue box) selected by our algorithm and the deblurred result.(f) The kernel
estimation and deblurred result using the whole image.(Best viewed on a high-resolution display).}\label{car}
\end{figure*}

Figure \ref{tower} - \ref{saturated} show the comparison results using
the patch selection algorithms of \cite{Fergus2006}, \cite{Xu2010} and \cite{HuZhe2015} respectively.
The deblurring results of the proposed method are better than other three region selection approaches.

In Figure \ref{text}, our result contains less ringing artifacts than others.
In Figure \ref{car} and \ref{84}, our details (shown in green boxes) are also the clearest of all the results.
One can find that although the selected patches are similar (e.g. Figure \ref{car} (b),(c) and (e)),  the recovered image on
region inferred by the proposed method is better than the other two.

\subsection{Comparison with Deblurring Using the Whole Image}
\begin{figure*}[!t]
  \hspace{-0.25cm}
  \includegraphics[width=1.00\linewidth]{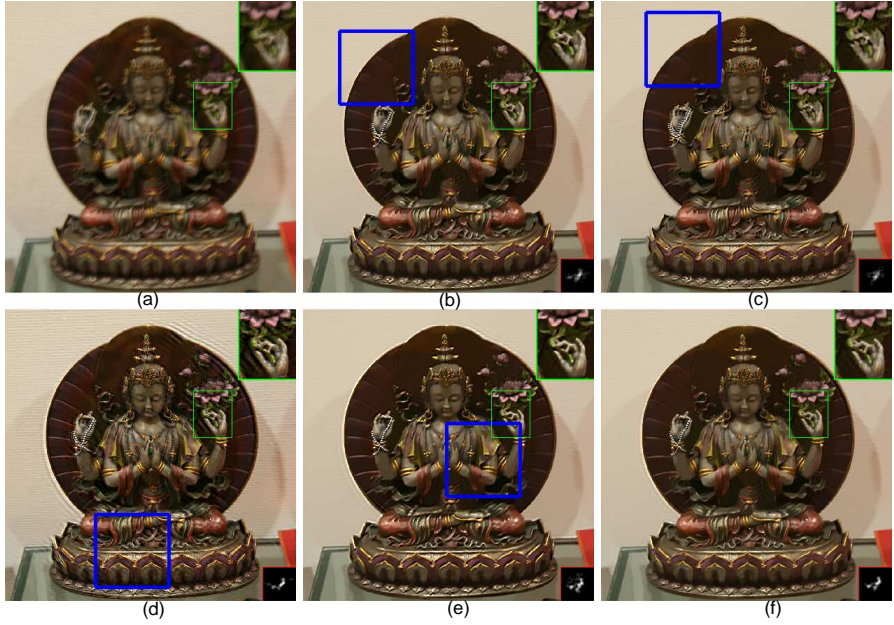}\\
  \vspace{-0.5cm}
  \caption{Inferred top sub-windows for a real blurry image and recovered results. We use the method
in \cite{Pan2016} for kernel estimation and  non-blind deconvolution. (a) Blurred image. (b)The subwindow (blue box) selected by window selector \cite{Fergus2006} and the deblurred result.  (c)The subwindow (blue box) selected by window selector \cite{Xu2010} and the deblurred result.(d)The subwindow (blue box) selected by window selector \cite{HuZhe2015} and the deblurred result.(e)The subwindow (blue box) selected by our algorithm and the deblurred result.(f) The kernel
estimation and deblurred result using the whole image. (Best viewed on a high-resolution display).}\label{84}
\end{figure*}

\begin{figure*}[!t]
  \hspace{-0.75cm}
  \includegraphics[width=1.1\linewidth]{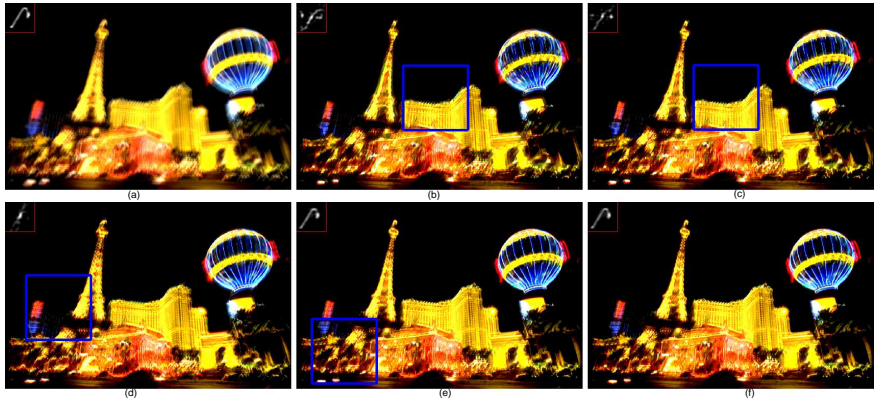}\\
  \vspace{-1.0cm}
  \caption{Inferred top sub-windows for a saturated image deblurring. We use the method
in \cite{Pan2016} for kernel estimation and  non-blind deconvolution. (a) Blurred image. (b)The subwindow (blue box) selected by window selector \cite{Fergus2006} and the deblurred result.  (c)The subwindow (blue box) selected by window selector \cite{Xu2010} and the deblurred result.(d)The subwindow (blue box) selected by window selector \cite{HuZhe2015} and the deblurred result.(e)The subwindow (blue box) selected by our algorithm and the deblurred result.(f) The kernel
estimation and deblurred result using the whole image. (Best viewed on a high-resolution display).}\label{saturated}
\end{figure*}

We also compare the algorithm results with \cite{Pan2016} which use the whole image to estimate blur kernels.  In this experiment, we select the top ranked patch to estimate blur kernels.
In Figure \ref{car} - \ref{saturated}, we show the comparison results of the algorithms.
Compared with the delurred results with the whole images, our method produces comparable or better kernel estimation and recovered images.

In addition, from Figure \ref{saturated}, we find that our method performs well for saturated images
even though the training set does not include such images.
Since the contrast between saturated and unsaturated regions, the saturated image contains lots of significant edges,
which are different from our training samples, but our model also can infer good regions for estimating the blur kernel.
This example shows that our network has good generalization ability.

\subsection{Quantitative Comparison}
\begin{figure}[!t]
  \includegraphics[width=1.1\linewidth]{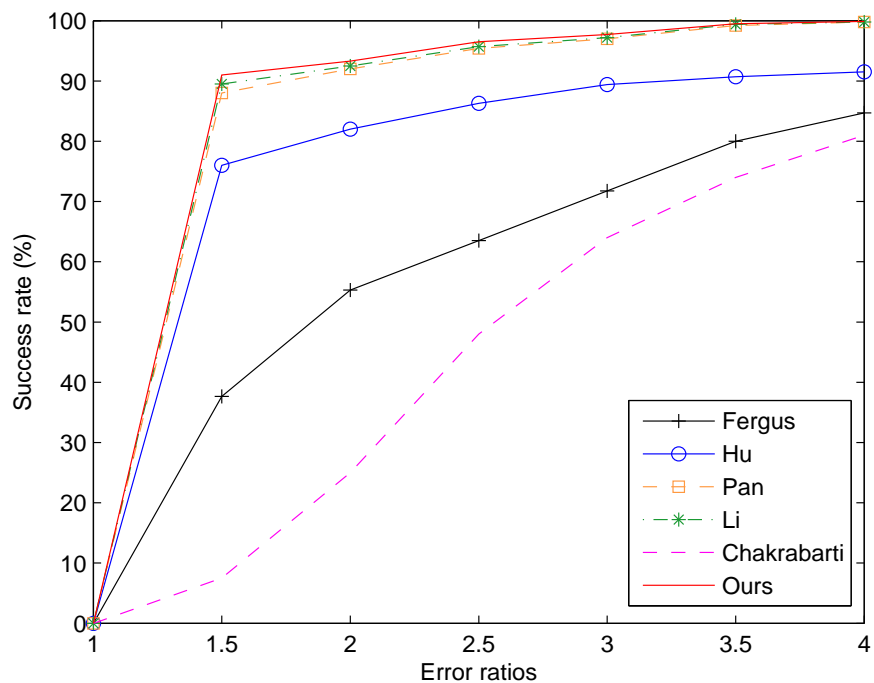}\\
  \vspace{-0.4cm}
  \caption{Success rate of reconstructed error ratio.}\label{curves}
\end{figure}

We use 640 challenging test images for extensive comparison, and demonstrate their cumulative error histograms.
The dataset was provided by Sun  $et~al.$\cite{Sun2013}, which consists of 80 sharp images and 8 blur kernels from Levin $et~al.$ \cite{Levin2009}. Given a good region by our method, we use the effective approach \cite{Pan2016} to estimate the blur kernel and recover the sharp image.
In order to conduct a comprehensive evaluation, we compare the results of the above region selection strategies \cite{Fergus2006,HuZhe2015}, and also use the method proposed by Pan $et~al.$\cite{Pan2016} to obtain the kernel and the clear image.
In addition, we compare our method with \cite{Pan2016}, \cite{Li2018} and \cite{Chakrabarti2016}  using the whole image
to estimate the blur kernel. We measure the error ratio \cite{Levin2009} and plot the results in Figure \ref{curves}.

From Figure \ref{curves}, one can find that the curve using our algorithm generally performs better than other approaches.
The results show that our region selection method is effective and robust, and that not all the features in the blurred image are useful for the kernel estimation. The whole image is used to estimate the blur kernel may not be the best choice.

\begin{figure}[!t]
  \includegraphics[width=1.0\linewidth]{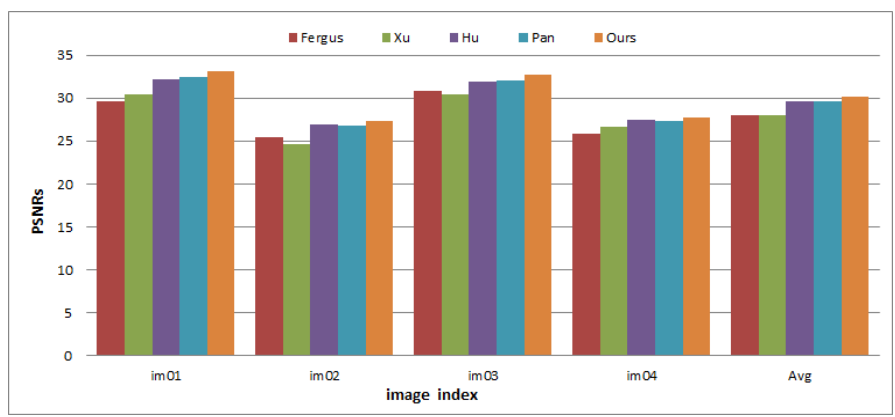}\\
  \vspace{-0.0cm}
  \caption{Quantitative comparison on the dataset \cite{Kohler2012} in terms of PSNR. The numbers below the horizontal axis denote the image index, the average PSNR values of all the images are shown on the rightmost column.}\label{bar}
\end{figure}

We also evaluate our method on the natural image dataset \cite{Kohler2012},
which contains 4 sharp images and 12 blur kernels.
We compare our method  with \cite{Pan2016} and the other three region selection approaches \cite{Fergus2006,Xu2010,HuZhe2015}. We compute the PSNR by comparing each
deblurred image with the clear image captured along the
same camera motion trajectory. As shown in Figure \ref{bar},
the proposed algorithm obtains the highest PSNR on average.

\section{Conclusions}
Recent image deblurring works focus on introducing image priors or end-to-end deep networks, and the research on the structure information of the blurred image is relatively less.
In this work, we study on the structural feature of blurred images and  exploit  them to address the problem of image deblurring.
We present a method based on deep learning to select good features and image regions for estimating the blur kernel.

Our proposed method can automatically choose appropriate image patches for deblurring, which has achieved good results.
It can avoid the tedious trials when users select image regions, and reduce the complexity of blur kernel estimation.



\begin{thebibliography}{}
%
%
\bibitem{Levin2009}
A.Levin,Y.Weiss,F.Durand,W.T.Freeman, ``Understanding and evaluating blind deconvolution algorithms,'' \emph{In Proceedings
of IEEE conference on computer vision and pattern recognition}, 1964-1971,2009.
\bibitem{Fergus2006}
R.Fergus, B.Singh, A.Hertzmann, Sam T. Roweis, and William T. Freeman, ``Removing camera shake
from a single photograph,'' \emph{ACM Transactions on Graphics}, vol. 25, no. 3, pp. 787--794, 2006.
\bibitem{Xu2010}
L.Xu and J.Jia, ``Two-phase kernel estimation for robust motion deblurring,'' \emph{In European Conference on Computer Vision}, 2010.
\bibitem{Joshi2008}
N.Joshi, R.Szeliski, D.J. Kriegman, ``PSF estimation using sharp edge prediction,'' \emph{
In Proceedings of IEEE conference on computer vision and pattern recognition}, 1-8,2008.
\bibitem{Cho2010}
T.S.Cho, N.Joshi, C.L. Zitnick, S.B.Kang, R.Szeliski, W.T.Freeman ``A content-aware image prior,'' \emph{In Proceedings of IEEE conference on computer vision and pattern recognition}, 169-176,2010.
\bibitem{Shan2008}
Q.Shan, J.Jia,  A.Agarwala, ``High-quality motion deblurring from a single
image,'' \emph{In: ACM SIGGRAPH},73:1-73:10, 2008.
\bibitem{Cho2009}
S. Cho and S. Lee, ``Fast motion deblurring,'' \emph{ ACM Transactions on Graphics}, vol. 28, no. 5, pp:145, 2009.
\bibitem{HuZhe2015}
Z.Hu, M.H .Yang, ``Learning Good Regions to Deblur Images,'' \emph{International Journal of Computer Vision}, vol. 115, no. 3,2015.
\bibitem{Lai2015}
W.S.Lai, J.J.Ding, Y.Y.Lin, and Y.Y.Chuang, ``Blur kernel estimation using normalized color-line
priors,'' \emph{In IEEE Conference on Computer Vision and Pattern Recognition}, 1-8, 2015.
\bibitem{Pan2014}
J. Pan, Z. Hu, Z. Su, and M. H. Yang. Deblurring text images via l0-regularized intensity and gradient prior. In IEEE Conference on Computer Vision and Pattern Recognition, 2014.
\bibitem{Pan2016}
J.Pan, D.Sun, H.Pfister, and M.H. Yang, ``Blind image deblurring using dark channel
prior,'' \emph{In IEEE Conference on Computer Vision and Pattern Recognition}, 1-8, 2016.
\bibitem{Yan2017}
Y.Yan, W.Ren, Y.Guo, R.Wang, and X.Cao, ``Image deblurring via extreme channels prior,'' \emph{
In IEEE Conference on Computer Vision and Pattern Recognition},1-8, 2017.
\bibitem{Chen2019}
L.Chen, F.Fang, T.Wang, ``Blind Image Deblurring With Local Maximum Gradient Prior,'' \emph{In IEEE Conference on Computer Vision and Pattern Recognition}, 1-8, 2019.

\bibitem{Schuler2016}
C. J. Schuler, M. Hirsch, S. Harmeling, and B. Scholkopf, ``Learning to deblur,'' \emph{IEEE Transactions on Pattern Analysis and Machine Intelligence}, vol. 38,no. 7, pp.1439--1451, 2016.

\bibitem{Hradis2015}
M. Hradis, J. Kotera, P. Zemck, and F.Sroubek, ``Convolutional neural networks for direct text deblurring,'' \emph{In British Machine Vision Conference}, 2015.
\bibitem{Yan2016}
R. Yan and L. Shao, ``Blind image blur estimation via deep learning,'' \emph{IEEE Transactions on Image Processing}, vol. 25, no.4,pp.1910--1921, 2016.
\bibitem{Yang2019}
H.Yang, Z.B.Zhang, Y.J.Guan, ``Rolling bilateral filter-based text image deblurring,'' \emph{The Visual Computer},vol. 35,no. 11, pp.1627--1640, 2019.

\bibitem{Kupyn2018}
O. Kupyn, V. Budzan, M. Mykhailych, D. Mishkin and J. Matas, ``DeblurGAN: Blind Motion Deblurring Using Conditional Adversarial Networks,'' \emph{In IEEE Conference on Computer Vision and Pattern Recognition}, Salt Lake City, UT, pp.8183-8192, 2018.

\bibitem{Kupyn2019}
O. Kupyn, T. Martyniuk, J. Wu and Z. Wang, ``DeblurGAN-v2: Deblurring (Orders-of-Magnitude) Faster and Better,'' \emph{In IEEE International Conference on Computer Vision}, Seoul, Korea (South), pp. 8877-8886, 2019.

\bibitem{Hosseini2020}
M. S. Hosseini and K. N. Plataniotis, ``Convolutional Deblurring for Natural Imaging,'' \emph{IEEE Transactions on Image Processing}, vol. 29, no. 1, pp. 250-264, 2020.

\bibitem{Bae2012}
H.Bae, C. C.Fowlkes, and  P. H.Chou, ``Patch mosaic for fast
motion deblurring,'' \emph{In Proceedings of Asian conference on computer vision},322-335,2012.

\bibitem{He2016}
K.M.He, X.Zhang,S.Ren, J.Sun, ``Deep Residual Learning for Image Recognition,'' \emph{In IEEE Conference on Coputer Vision  and Pattern Recognition}, 1-8, 2016.

\bibitem{Li2018}
L. Li, J. Pan, W.S. Lai, C. Gao, N.Sang, and M.H. Yang, ``Learning a discriminative prior for blind image deblurring,'' \emph{In Proceedings of IEEE conference on computer vision and pattern recognition}, 6616-6625, 2018.

\bibitem{Chakrabarti2016}
A. Chakrabarti, ``A neural approach to blind motion deblurring,'' \emph{In European Conference on Computer Vision},  221-235, 2016.
\bibitem{Sun2013}
L. Sun, S. Cho, J. Wang, and J. Hays, ``Edge-based blur kernel estimation using patch priors,'' \emph{In IEEE International Conference on Computational Photography}, 1-8, 2013.

\bibitem{Kohler2012}
R. K\" ohler, M. Hirsch, B. Mohler, B. Sch\" olkopf, and S. Harmeling, ``Recording and playback of camera
shake: Benchmarking blind deconvolution with a real-world database,'' \emph{In European Conference on Computer Vision},
27-40, 2012.
\end{thebibliography}


\end{document}